\documentclass[final]{cvpr}

\usepackage{times}
\usepackage{epsfig}
\usepackage{graphicx}
\usepackage{amsmath}
\usepackage{amssymb}
\usepackage{makecell}
\usepackage{booktabs}
\usepackage{multirow}
\usepackage{pifont}
\newcommand{\cmark}{\ding{51}}%
\newcommand{\xmark}{\ding{55}}%
\usepackage[pagebackref=true,breaklinks=true,colorlinks,bookmarks=false]{hyperref}

\newcommand{\tabincell}[2]{\begin{tabular}{@{}#1@{}}#2\end{tabular}}  
\def\cvprPaperID{708} % *** Enter the CVPR Paper ID here
\def\confYear{CVPR 2021}

\begin{document}

%%%%%%%%% TITLE
\title{Temporal Context Aggregation Network for Temporal Action \\ Proposal Refinement
\vspace{-0.4cm}
}

% \author{Zhiwu Qing$^{1\ast}$\quad 
% Haisheng Su$^{2\dag}$ \quad
% Weihao Gan$^{2}$ \quad
% Dongliang Wang$^{2}$ \quad
% Wei Wu$^{2}$ \and 
% Xiang Wang$^{1}$\quad
% Yu Qiao$^{3,4}$\quad
% Junjie Yan$^{2}$\quad
% Changxin Gao$^{1}$\quad
% Nong Sang$^{1\dag}$ \\
% $^{1}$Key Laboratory of Image Processing and Intelligent Control, \\
% School of Artificial Intelligence and Automation, Huazhong University of Science and Technology \\ $^{2}${SenseTime Research} \\ $^{3}$Shenzhen Institutes of Advanced Technology, Chinese Academy of Sciences\\
% $^{4}$Shanghai AI Laboratory, Shanghai, China\\
% {\tt\small \{qzw,wxiang,cgao,nsang\}@hust.edu.cn, \tt\small yu.qiao@siat.ac.cn \\ \tt\small \{suhaisheng,ganweihao,wangdongliang,wuwei,yanjunjie\}@sensetime.com}
% % just like the second author
% \vspace{-0.6cm}
% }
\author{Zhiwu Qing$^{1\ast}$\quad 
Haisheng Su$^{2\dag}$ \quad
Weihao Gan$^{2}$ \quad
Dongliang Wang$^{2}$ \quad
Wei Wu$^{2}$ \and 
Xiang Wang$^{1}$\quad
Yu Qiao$^{3,4}$\quad
Junjie Yan$^{2}$\quad
Changxin Gao$^{1}$\quad
Nong Sang$^{1\dag}$ \\
$^{1}$Key Laboratory of Image Processing and Intelligent Control, \\
School of Artificial Intelligence and Automation, Huazhong University of Science and Technology \\ $^{2}${SenseTime Research} \\ $^{3}$Shenzhen Institutes of Advanced Technology, Chinese Academy of Sciences\\
$^{4}$Shanghai AI Laboratory, Shanghai, China\\
{\tt\small \{qzw,wxiang,cgao,nsang\}@hust.edu.cn, yu.qiao@siat.ac.cn }\\
{\tt\small\{suhaisheng,ganweihao,wangdongliang,wuwei,yanjunjie\}@sensetime.com}
\vspace{-0.4cm}
}

\maketitle
\pagestyle{empty}  % no page number for the second and the later pages
\thispagestyle{empty} % no page number for the first page
\let\thefootnote\relax\footnotetext{$\ast$ The work was done during an internship at SenseTime.}
\let\thefootnote\relax\footnotetext{$\dag$ Corresponding author.}
%%%%%%%%% ABSTRACT
\begin{abstract}
\vspace{-0.3cm}
Temporal action proposal generation aims to estimate temporal intervals of actions in untrimmed videos, which is a challenging yet important task in the video understanding field.
The proposals generated by current methods still suffer from inaccurate temporal boundaries and inferior confidence used for retrieval owing to the lack of efficient temporal modeling and effective boundary context utilization.
In this paper, we propose Temporal Context Aggregation Network (TCANet) to generate high-quality action proposals through ``local and global" temporal context aggregation and complementary as well as progressive boundary refinement.
Specifically, we first design a Local-Global Temporal Encoder (LGTE), which adopts the channel grouping strategy to efficiently encode both ``local and global" temporal inter-dependencies.
Furthermore, both the boundary and internal context of proposals are adopted for frame-level and segment-level boundary regressions, respectively.
Temporal Boundary Regressor (TBR) is designed to combine these two regression granularities in an end-to-end fashion, which achieves the precise boundaries and reliable confidence of proposals through progressive refinement. Extensive experiments are conducted on three challenging datasets: HACS, ActivityNet-v1.3, and THUMOS-14, where TCANet can generate proposals with high precision and recall. By combining with the existing action classifier, TCANet can obtain remarkable temporal action detection performance compared with other methods. Not surprisingly, the proposed TCANet won the 1$^{st}$ place in the CVPR 2020 - HACS challenge leaderboard on temporal action localization task.
\end{abstract}

\begin{figure}[t]
\centering
\includegraphics[width=8.cm, height=5cm]{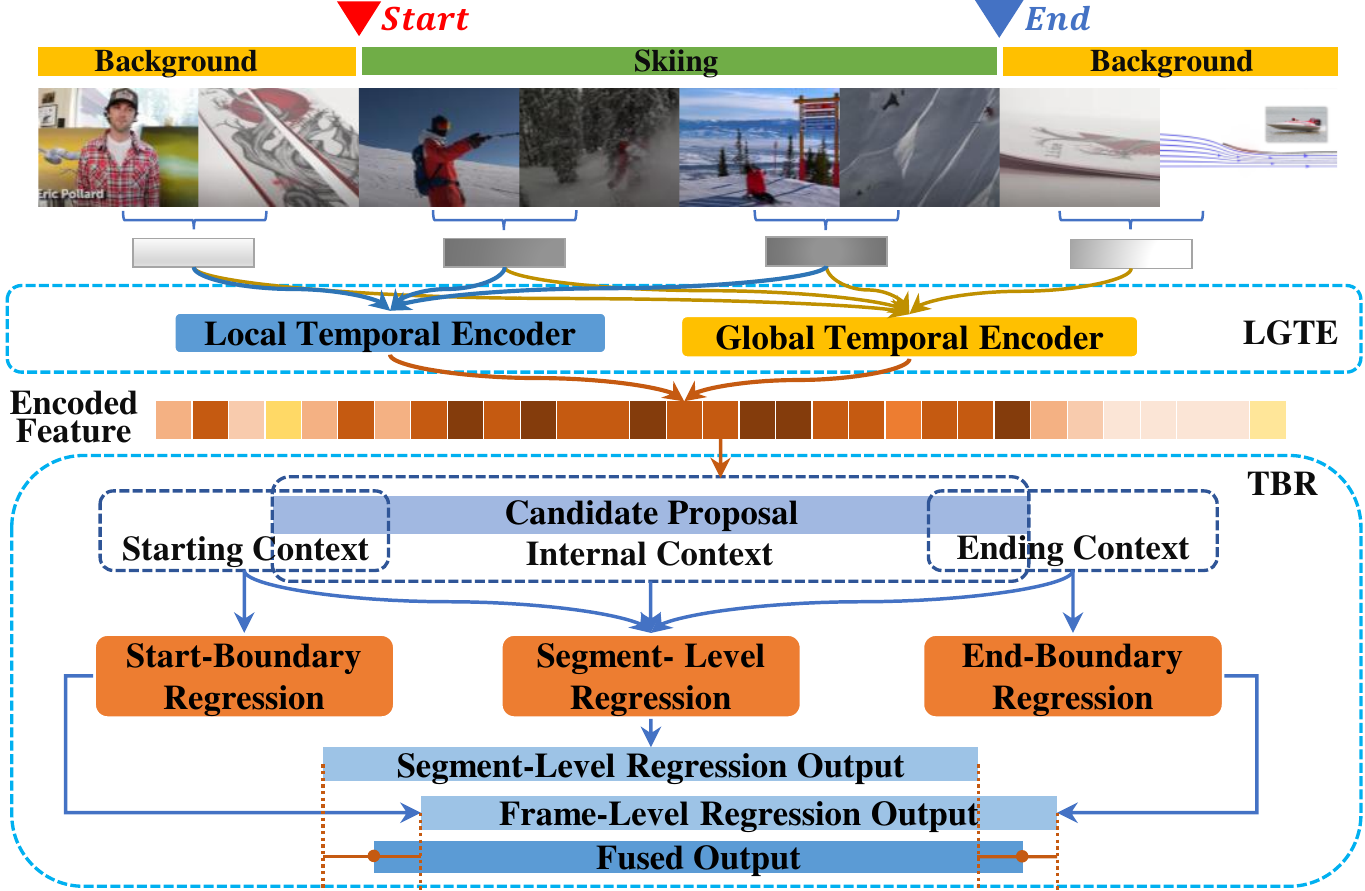}
\caption{Overview of our proposed method. Given an untrimmed video, TCANet captures the ``local and global" temporal relationships in parallel by LGTE.
In TBR, the internal and boundary context of proposals are utilized for segment-level boundary regression and frame-level boundary regression, respectively.
}
\label{fig_overview}
\vspace{-0.2cm}
\end{figure}

%%%%%%%%% BODY TEXT
\section{Introduction}
The temporal action detection task requires locating the starting and ending time of action instances from long untrimmed videos and classifying the actions.
This task can be applied to many fields, such as video content analysis and video recommendation.
Most existing temporal action detection methods follow a two-stage scheme~\cite{wang2015ssn,singh2016anet_winner}, namely temporal action proposal generation and classification.
Although action recognition methods~\cite{wang2016tsn,feichtenhofer2019slowfast} have achieved impressive classification accuracy, 
the temporal action detection performance is still unsatisfactory in several mainstream benchmarks~\cite{thumos14, caba2015activitynet, zhao2019hacs}.
Hence, many researchers target improving the quality of temporal action proposals.

Current proposal generation methods can be mainly divided into two steps.
First, the temporal relationship is captured by stacked temporal convolutions ~\cite{lin2018bsn, lin2019bmn, lin2020dbg,su2020transferable,su2018cascaded} or global temporal pooling operations~\cite{gao2020rapnet}. Then proposals are further generated by the boundary-based regression methods~\cite{lin2018bsn,lin2019bmn} or the anchor-based regression methods~\cite{shou2016scnn,gao2017turn,gao2018ctap,chao2018tal_net,liu2019mgg,gao2020rapnet}.
However, existing methods share the following drawbacks:
(1) Neither convolution nor global fusion can effectively model the temporal relationship. The 1$D$ convolution operations~\cite{lin2018bsn, lin2019bmn, lin2020dbg} lack flexibility in encoding long-term temporal relationships limited by kernel size.
The global fusion methods~\cite{gao2020rapnet} neglect various global dependencies for each temporal location and the implicit attention to local details, such as local details of boundaries. Besides, simply collecting global features through average pooling may introduce unnecessary noise.
(2) 
{\color{black} Only the internal context or boundary context of proposals used for regression is inferior to generate proposals with precise boundaries.
The \textit{internal context} of proposals adopted in anchor-based methods can obtain reliable confidence scores but fails to generate precise boundaries. On the contrary,  the \textit{boundary context} of proposals considered in boundary-based methods is sensitive to boundary changes but generates proposals with inferior proposal-level confidence.}
%without proposal-level features

%
To relieve these issues, we propose Temporal Context Aggregation Network (TCANet) for high-quality proposal generation, as shown in Figure~\ref{fig_overview}.
%from two main perspectives.
%
First, the Local-Global Temporal Encoder (LGTE) is proposed to simultaneously capture \textit{local and global} temporal relationships in a channel grouping fashion, which contains two main sub-modules.
Specifically, the input features after linear transformation are equally divided into $N$ groups along the channel dimension. Then Local Temporal Encoder (LTE) is designed to handle the first $A$ groups for local temporal modeling. At the same time, the remaining $N - A$ groups are captured by the Global Temporal Encoder (GTE) for global information perception.
%are projected , and then the channels are equally divided into $N$ groups.
%
%The first $A$ groups are only designed to capture the local temporal relationships, which we call Local Temporal Encoders (LTEs). 
%While the remaining $N-A$ groups are responsible for modeling the global inter-dependencies, called Global Temporal Encoders (GTEs). 
%
In this way, LGTE is expected to integrate the long-term context of proposals by global groups while recovering more structure and detailed information by local groups. 
Second, the Temporal Boundary Regressor (TBR) is proposed to exploit both boundary context and internal context of proposals for frame-level and segment-level boundary regressions, respectively. Concretely, the frame-level boundary regression aims to refine the starting and ending locations of candidate proposals with boundary sensitivity, while the segment-level boundary regression aims to refine the center location and duration of proposals under the overall perception of proposals. Finally, high-quality proposals are obtained through complementary fusion and progressive boundary refinements.
%Concretely, boundary context is utilized to refine the starting and ending locations of proposals because of their sensitivity to temporal boundaries. internal context is adopted for segment-level regression under the overall perception of proposals.
%regress center and width because they have the overall perception of proposals. The complementary fusion of these two contexts can be beneficial for proposal boundary quality improvement.
%
%Finally, three cascaded TBRs are further employed to achieve better proposal boundaries from coarse to fine.

In summary, our contributions mainly have three folds:

\begin{itemize}
\item {We design a Local-Global Temporal Encoder to simultaneously capture \textit{local and global} temporal relationships in a channel grouping fashion. It can be easily embedded into any other proposal generation frameworks for efficient temporal relationships modeling.}
\item {Temporal Boundary Regressor is proposed to perform complementary and progressive boundary refinements, including the local frame-level boundary regression and global segment-level boundary regression.
% A multi-stage refinement strategy is further adopted to improve the proposal boundaries and confidence score from coarse to fine.
}
\item {Extensive experiments reveal that TCANet can obtain convincing proposals performance on several benchmarks: HACS, ActivityNet-v1.3, and THUMOS-14. By combining with the existing classifier, TCANet can achieve remarkable temporal action detection performance compared with other approaches.}
\end{itemize}

\begin{figure*}[t]
\centering
\includegraphics[width=14.7cm, height=2.5cm]{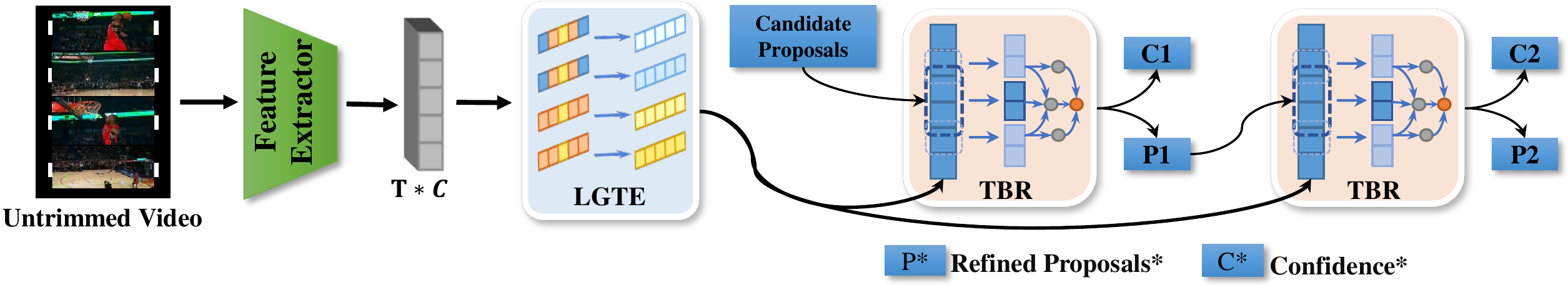}
\caption{{\color{black}The framework of TCANet. TCANet mainly contains two modules: LGTE and TBR.
LGTE is employed to capture \textit{local-global} temporal inter-dependencies simultaneously. TBR is adopted to perform frame-level and segment-level boundary regressions, respectively.
Finally, high-quality proposals are obtained through complementary fusion and progressive boundary refinements.
}
}
\label{fig_framework}
\vspace{-0.2cm}
\end{figure*}

\section{Related Work}
\noindent \textbf{Action Recognition.}
Action recognition is an important task in video understanding area, which is in need of spatio-temporal information modeling.
Current deep learning-based action recognition methods can be mainly divided into three types. 
The first type is 2stream networks~\cite{simonyan2014twostream,wang2015twostream2,feichtenhofer2016twostream1}, which adopt RGB frames and optical flow to capture appearance and motion information. The second type ~\cite{tran2015learning_c3d,carreira2017quo_inception_3d,xie1712rethinking,tran2018closer_look} directly captures spatio-temporal information from raw videos using 3D convolution.
The third type aims to efficiently model spatio-temporal features by decoupled (2 + 1) D convolutions~\cite{qiu2017learning_p3d,feichtenhofer2019slowfast,lin2019tsm,jiang2019stm,li2020tea,su2020collaborative}.
In this work, 2stream and SlowFast are adopted to encode the input video features.

\noindent \textbf{Temporal Action Proposal Generation and Detection.}
Current proposal generation methods can be mainly divided into \textit{anchor-based} and \textit{boundary-based} methods. The \textit{anchor-based} methods~\cite{wang2015ssn,shou2016scnn,gao2017turn,gao2018ctap,liu2019mgg} refer to the temporal boundary refinements of sliding windows or pre-defined anchors. Among them, TURN~\cite{gao2017turn} and CTAP~\cite{gao2018ctap} directly concatenate the boundary context and internal context of proposals for boundary refinements (i.e., starting and ending locations), while other methods aim to refine the duration and center location of proposals.
%refer to the usage of sliding windows or pre-defined anchors to initialize candidate proposals, then refine the temporal boundaries of proposals. 
However, refinements on boundary locations only cannot make full use of contextual information of proposals, while mere refinement of duration and center location of candidate proposals would also neglect the local boundary details. Therefore, it is non-trivial to combine these two regression granularities into a unified framework.
%However, local boundary context will lose contextual information of proposals and be easier to generate noise boundaries, while global proposal context lack the sensitivity of boundaries. 
% However, global proposal context can int
%
\textit{Boundary-based} methods~\cite{lin2018bsn,lin2019bmn,su2020bsn++} first generate the boundary probability sequence, then apply the Boundary Matching mechanism to generate candidate proposals. MGG~\cite{liu2019mgg} simply combines a boundary-based stream and anchor-based stream with a shared backbone to extract features, then each stream is optimized independently, and the results are fused during the inference.
In our work, we make full use of boundary context and internal context of proposals to predict the frame-level offsets (i.e., starting and ending) and the segment-level offsets (i.e., center and duration), respectively. Meanwhile, we jointly train these two granularities with supervision performed on the combined results. Finally, complementary and progressive boundary refinements are conducted for better performance. 

% Although these methods can generate proposals with high recall, it still suffer from inaccurate boundaries and inferior confidence used for retrieval.
%In our work, we leverage the complementarity between boundary context and internal context of proposals for frame-level boundary regression and segment-level boundary regression, and adopt a progressive strategy to refine proposals from coarse to fine.

\noindent \textbf{Self-Attention Mechanism.}
The self-attention~\cite{wang2018nonlocal} mechanism is widely used in the video understanding area since it can effectively capture long-term dependencies compared with other attention methods such as recurrent models~\cite{sharma2015rnn} and pooling methods~\cite{girdhar2017attentionalpooling}.
The Transformer~\cite{vaswani2017transformer} is also based on the self-attention mechanism, which is originally applied in the machine translation task. Girdhar \etal ~\cite{girdhar2019ava_transformer,gavrilyuk2020actor_transformer} adopt Transformer to capture the interactions between human and objects existing in videos.
In this paper, we propose Local-Temporal Global Encoder, which can efficiently capture both ``local and global" temporal relationships and then integrate rich contexts into extracted video features for temporal proposals generation.

\section{TCANet}
As shown in Figure~\ref{fig_framework}, we propose \textbf{Temporal Context Aggregation Network (TCANet)} to generate high-quality proposals, which mainly consists of two main modules: Local-Global Temporal Encoder and Temporal Boundary Regressor.
Firstly, the Local-Global Temporal Encoder (LGTE) is adopted to simultaneously encode the input video features' \textit{local and global} temporal relationships. {\color{black} Then the Temporal Boundary Regressor (TBR) is utilized to refine the boundaries of the proposals by exploiting both boundary and internal context for frame-level and segment-level boundary regressions, respectively.}

\subsection{Problem Formulation}
For an untrimmed video $X$ with $l$ frames, we can denote it as $X=\{x_i\}_{i=1}^{l}$, where $x_i$ is the $i$-th frame of the video. Temporal proposal generation task is to generate a set of proposals $P=\{t_{sj},t_{ej}\}_{j=1}^{N_p}$ that may contain action instances for video $X$, where $t_{sj}$ and $t_{ej}$ are the starting time and ending time of the $j$-th proposal, and $N_p$ is the number of proposals.

\subsection{Feature Encoding}
For a given video, features are extracted by SlowFast~\cite{feichtenhofer2019slowfast} and 2stream~\cite{simonyan2014twostream} since their excellent performance on the video classification task. The frame rate of videos is set to $r$ fps, and each snippet contains $s$ frames. Each snippet is encoded into a visual feature $f_i\in R^C$ by a feature extractor. Given an untrimmed video, a video feature sequence of $F={\{f_i\}}_{i=1}^T\in R^{T\times C}$ is obtained by this method, where $T=l/\delta$, $l$ is the total number of video frames, and $\delta$ is the number of frames interval between different snippets.

%by exploiting the complementarity of local boundary features and global central features of proposals.}

\begin{figure}[t]
\centering
\includegraphics[width=8.2cm,height=6.0cm]{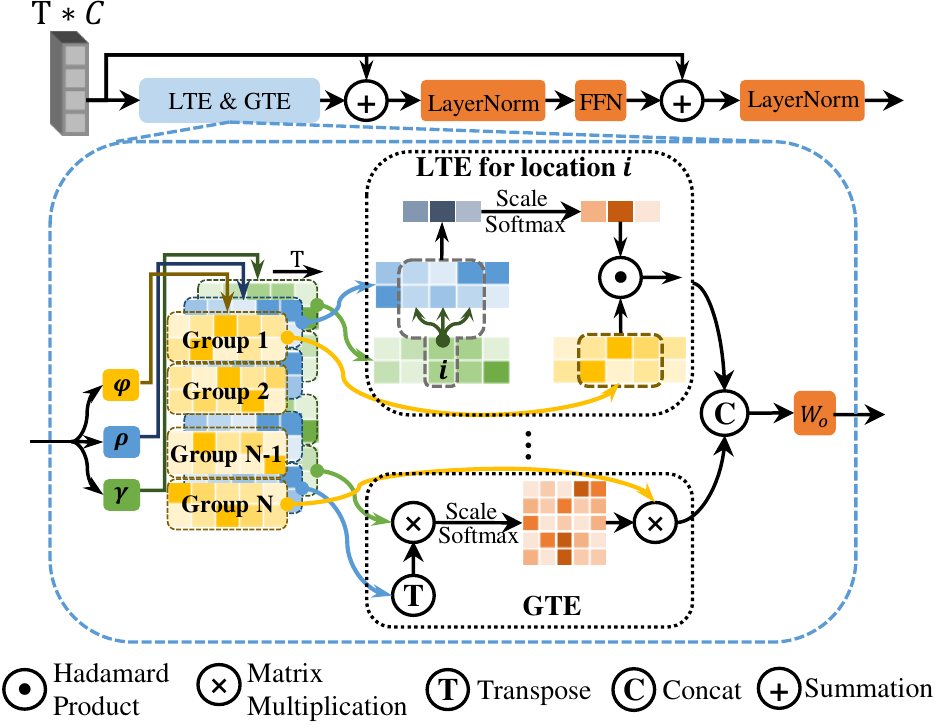}
\caption{The detailed structure of Local-Global Temporal Encoder (LGTE). The input features are divided into $N$ groups along the channel dimension. 
Then the first $A$ groups are fed to $A$ Local Temporal Encoders (LTE) separately, where the local dynamic modeling is achieved by calculating the regional attention for each temporal location.
The remaining $(N-A)$ groups are adopted by $(N-A)$ Global Temporal Encoders (GTE) separately to calculate the similarity between each location and global feature sequence.
$W_{o}$ is a learnable matrix, and Feed Forward Network (FFN) is a nonlinear projection function.
}
\label{fig_lgte} 
\vspace{-0.2cm}
\end{figure}

\subsection{Local-Global Temporal Encoder}
For long videos, long-term temporal dependency modeling is essential, proven by many previous works~\cite{gao2020rapnet,wu2019lfb}. 
Nonlocal~\cite{wang2018nonlocal} is often applied to obtain the relationship between different global locations.
However, global modeling only is easy to introduce global noise and insensitive to small boundary changes. 
We propose a local and global joint modeling strategy to alleviate this problem, as shown in Figure~\ref{fig_lgte}.

\noindent \textbf{Local Temporal Encoder (LTE)} is responsible for capturing local dependencies based on local details dynamically. Precisely, to measure the relationship between temporal location $i$ and its local areas, the $cosine$ similarity between two temporal locations is adopted to generate  similarity vector $S_i^l$ and weight vector $W_i^l$:
\begin{equation}
    S_i^l=\gamma^l(f_{i})\cdot(\rho^l([(f_{i-\lfloor w/2 \rfloor})^T,\cdot \cdot \cdot (f_{i+\lfloor w/2 \rfloor})^T]^T))^T \in R^{1\times w}
\label{local_sim}
\end{equation}
\begin{equation}
    W_i^l = \text{Softmax}(\frac{S_i^l}{\sqrt{C}})
\label{softmax_fuse1}
\end{equation}
where $C$ is the number of channels, $w$ is the size of the modeling area for location $i$, which is defined as $WindowSize$. For example, the value of $w$ in Figure~\ref{fig_lgte} LTE is 3. $\gamma^l$ and $\rho^l$ are two different linear projection functions that map the input feature vectors to the similarity measure space.

With equation~\ref{local_sim}, the relationship between each location and its corresponding modeling area can be calculated. To achieve local information exchange, the following formula will be utilized to collect local context information from the corresponding local area dynamically:
\begin{equation}
    f_i^{l}=W_i^l \cdot(\varphi^l([(f_{i-\lfloor w/2 \rfloor})^T,\cdot \cdot \cdot (f_{i+\lfloor w/2 \rfloor})^T]^T)),
\label{local_fo}
\end{equation}
where $f_i^{l}$ represents the new expression of location $i$, and $\varphi^l$ is a linear projection function.

\noindent \textbf{Global Temporal Encoder (GTE)} is designed to model the long-term temporal dependencies of videos. Compared with LTE, GTE needs to aggregate global interactions for each location on the temporal dimension. Therefore, the relationship between each location and the global feature is written as follows:
\begin{equation}
    S_i^g=\gamma^g(f_{i})\cdot(\rho^g(F))^T \in R^{1\times T},
\label{global_sim}
\end{equation}
\begin{equation}
    W_i^g = \text{Softmax}(\frac{S_i^g}{\sqrt{C}}),
\label{softmax_fuse2}
\end{equation}
where $\gamma^g$ and $\rho^g$ are two different linear projection functions.
The global interaction feature of location $i$ can be updated by weight vector $W_i^g$:
\begin{equation}
    f_i^{g}=W_i^g \cdot(\varphi^g(F)),
\label{global_fo}
\end{equation}
where $f_i^{g}$ represents the new global feature representation of location $i$, and $\varphi^g$ is a linear projection function.

\noindent \textbf{Local-Global Temporal Encoder (LGTE).} Each location in the video feature sequence can be modeled locally and globally by LTE and GTE, respectively.
However, it is inefficient to combine them in the form of \textit{``LTE-GTE"} simply.
To solve this problem, LGTE is implemented in a channel grouping fashion.
Specifically, as shown in Figure~\ref{fig_lgte}, the input feature is first projected by $\gamma$, $\rho$, and $\varphi$. These outputs are then divided into $N$ groups along the channel dimension. Hence the channel number of each group is $C/N$. The first $A$ groups are handled by LTEs, while the other $N-A$ groups are fed to GTEs. For location $i$, the combined output of local and global features can be written as:
\begin{gather}
    f_{i}^{a}=[(f_{1i}^{l})^T, \cdot \cdot \cdot  (f_{Ai}^{l})^T,  (f_{(A+1)i}^{g})^T  \cdot \cdot \cdot  (f_{Ni}^{g})^T]^T\cdot W_{o},  \label{output_fuse}\\
    f_{i}^{b} = \text{LayerNorm}(f_{i}^{a}) + f_{i}^{a}, \\
    f_{i}^{'} = \text{LayerNorm}(\text{FFN}(f_{i}^{b}) + f_{i}^{b}),
\end{gather}
where $W_{o}$ is a learnable parameter matrix. Inspired by Transformer~\cite{vaswani2017transformer}, FFN is adopted to capture the interaction of features among different groups at $i$-th temporal location: $\text{FFN}(x)=\text{ReLu}(x\cdot W_1+b_1)\cdot W_2+b_2$.
% Equations~\ref{output_fuse} realize the interaction of features between different groups of the $i$-th temporal location.

\noindent \textbf{Discussion.} We notice that our LTE is similar to the convolution with fixed kernels. However, the dynamic local interaction modeling for each temporal location is unique for better adaptation to complex temporal changes than conventional convolution.
%local interaction is dynamic for each temporal locations and better adapts to complex temporal changes than convolution. 
Furthermore, the combination of LTE and GTE enables our LGTE to capture the global dependencies of whole videos and dynamically model local changes with less noise. Besides, the channel grouping fashion ensures high computing efficiency and the diversity of ``local and global" relationships.

\begin{figure}[t]
\centering
\includegraphics[width=6cm]{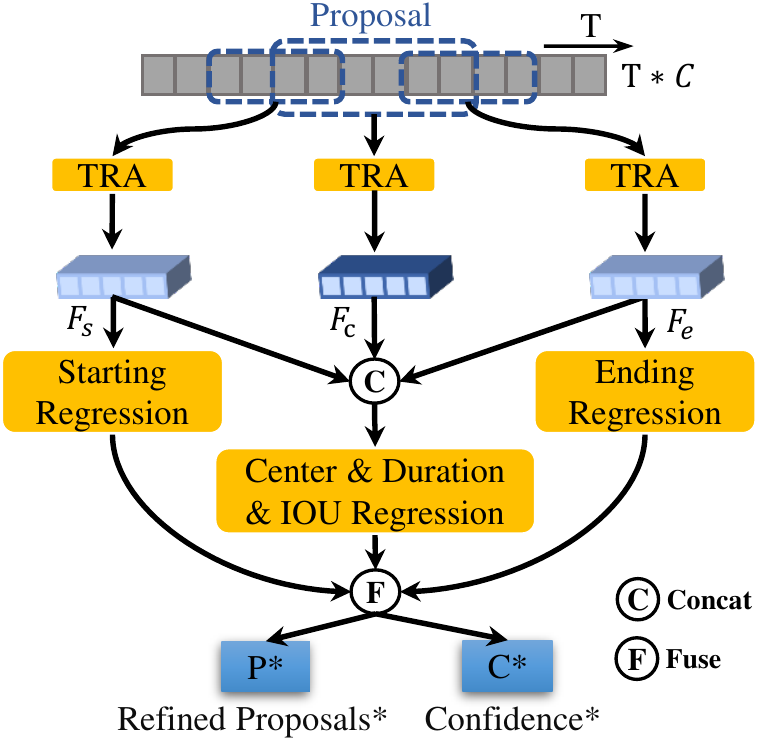}
\caption{The detailed structure of Temporal Boundary Regressor (TBR). For each input proposal, TBR collects its starting context  $F_s$, internal context $F_c$ and ending context $F_e$ through Temporal Roi Align (TRA). $F_s$ and $F_e$ are adopted to refine the starting and ending locations, respectively. $F_s$, $F_c$, and $F_e$ are concatenated to refine the center location and the duration of proposals. Finally, the accurate proposals are obtained by fusing these two outputs.
}
\label{fig_tbr} 
\vspace{-0.2cm}
\end{figure}

\noindent \subsection{Temporal Boundary Regressor}
\textit{Anchor-based methods} ~\cite{shou2016scnn,gao2017turn,lin2017ssad,gao2018ctap,chao2018tal_net,liu2019mgg,gao2020rapnet} leverage the internal context of proposals to regress center location and duration, which can obtain reliable scores but with lower recall.
\textit{Boundary-based methods}~\cite{lin2018bsn,lin2019bmn} only utilize local boundary context to locate the boundaries, which are sensitive to boundaries but with inferior confidence.
{\color{black} Therefore, we propose to combine the boundary context-based frame-level regression and the internal context-based segment-level regression to refine the boundaries.}

\noindent
\textbf{Complementary Regression Strategy.}
As shown in Figure~\ref{fig_tbr}, the feature of one proposal is divided into three parts: the starting context $F_s$, the internal context $F_c$, and the ending context $F_e$. 
To achieve frame-level regression, $F_s$ and $F_e$ are utilized to regress the boundary offset $\Delta \hat{s}$ and $\Delta \hat{e}$ of the starting time and ending time, respectively:
\begin{gather}
     \{\Delta\hat{s}, \Delta\hat{e}\}=\text{Conv1d}(\text{ReLu}(\text{Conv1d}(\{F_s, F_e\})))
\end{gather}
The boundary offsets $\Delta\hat{s}$ and $\Delta\hat{e}$ obtained by this method only utilize the starting and ending local features of proposals. It can effectively reduce noise interference and is more sensitive to the boundary position.

However, only using the local features of the boundary will lose the global context of proposals. Therefore, $F_s$, $F_c$ and $F_e$ are utilized to achieve segment-level regression, which jointly regress the center location offset $\Delta \hat{x}$ and duration offset $\Delta \hat{w}$ of the proposals:
\begin{gather}
    F_a=[F_s, F_c, F_e], \\ 
    \{\Delta \hat{x}, \Delta \hat{w}, p_{conf}\}=\text{Conv1d}(\text{ReLu} (\text{Conv1d}(F_a))), 
\end{gather}
By means of $\Delta\hat{s}$, $\Delta\hat{e}$, $\Delta\hat{x}$ and $\Delta\hat{w}$, two new proposals $(\hat{s_1}, \hat{e_1})$ and $(\hat{s_2}, \hat{e_2})$ can be obtained:
\begin{gather}
    \hat{s_1} = s_p - \Delta \hat{s} w_p, \quad  \hat{e_1} = e_p - \Delta \hat{e} w_p, \\
    \hat{x_2} = x_p - \Delta \hat{x} w_p, \quad  \hat{w_2} = w_p e^{\Delta \hat{w}}, \\
    \hat{s_2} = \hat{x_2} - \hat{w_2}/2, \quad \hat{e_2} = \hat{x_2} + \hat{w_2}/2,
\end{gather}
where $w_p=e_p-s_p$, denotes the length of the proposals.
Finally, the two new proposals will be fused as the final proposals prediction of TBR:
\begin{equation}
    \hat{s} = \tau\hat{s_1} + (1-\tau)\hat{s_2}, \quad \hat{e} = \tau\hat{e_1} + (1-\tau)\hat{e_2},
\end{equation}
where $\tau$ is a fusion parameter, we set it to 0.5 empirically.

% \textbf{Discussion.} TBR utilizes the complementarity of boundary context and internal context of proposals to realize the continuous refinement of boundaries. 
%

\noindent \textbf{Progressive Refinement.}
To achieve more accurate boundaries of candidate proposals, a progressive refinement strategy is adopted to generate high-quality proposals from coarse to fine. In ablation experiments, we will explore the impact of the number of TBRs on proposal performance.

%cascading multiple TBRs is adopted to obtain high quality proposals. 

\noindent \textbf{Discussion.} 
{\color{black} The boundary features based frame-level regression is sensitive to the local changes, which are helpful to detect the action boundaries caused by shot switching. And the internal proposal features based segment-level regression has an overall perception of proposals suitable for detecting actions with indistinct boundaries. Therefore, these two features are complementary and essential for generalized action proposal generation. MGG~\cite{liu2019mgg} proposes a multi-granularity generator to integrate boundary-based regression and anchor-based regression into a unified network with a shared backbone used for feature extraction. However, these two regression streams are trained independently, and the results are fused directly during inference. On the contrary, the main idea of TBR is to adopt both boundary and internal context of proposals to predict the frame-level and segment-level offsets, respectively, but jointly train these two granularities with supervision performed on the combined results for gradient backpropagating. Meanwhile, complementary and progressive boundary refinements are conducted for better performance.}

%simply combines boundary-based method (\textit{e.g.} BSN~\cite{lin2018bsn}) and anchor-based method (\textit{e.g.} SSAD~\cite{lin2017ssad}) during inference fusion.
%However, the motivation of TBR is to adopt the complementarity of boundary context  based frame-level regression and internal context based segment-level regression  online, which realize the joint and dynamic optimization of the two regression methods.}

%boundaries in continuous shots

\section{Training and Inference of TCANet}
\subsection{Training}
\noindent \textbf{Proposals Selection.}
To better demonstrate the effectiveness of TCANet, we adopt the proposals output from the most competitive method(BMN)~\cite{lin2019bmn} as our pool of candidates. 
In the training phase of TCANet, to make the network learning more efficient, Soft-NMS~\cite{bodla2017softnms} is adopted to preprocess proposals output by BMN to reduce the redundant samples.
Then the top 100 proposals are selected in descending order for training. 

\noindent \textbf{Label Assignment.}
During the TBR training process, proposals with ground truth temporal Intersection-over-Union(tIoU) greater than a certain threshold $i_p$ are specified as positive samples, and ground truth tIoU less than a certain threshold $i_n$ as negative samples, and those between $i_n$ and $i_p$ Proposals are incomplete samples. The number of positive samples, negative samples and incomplete samples are defined as $N_{pos}$, $N_{incomp}$ and $N_{neg}$, respectively. 
Three kinds of samples are randomly sampled so that $N_{pos}:N_{incomp}:N_{neg}=1:1:1$ in training.

\noindent \textbf{Loss Function.}
The loss functions of the IoU prediction and the position regression of proposals are denoted as $L_{iou}$ and $L_{reg}$, respectively.
We denote $L_{iou}$ and $L_{reg}$  as:
\begin{small}
\begin{gather}
    L_{iou} = \frac{1}{N_{train}}\left(\sum_{i=1}^{N_{train}}\text{SmoothL1}(p_{conf,i}, g_{iou, i})\right), \\
    L_{reg} = \frac{1}{N_{pos}}\left(\sum_{i\in Pos}^{N_{pos}}\sum_{\substack{m\in\{x,w,\\s,e\}}}\text{SmoothL1}(r_i^m-g_i^m)\right),
\end{gather}
\end{small}
where
\begin{equation}
    \begin{split}
    N_{train} = N_{pos} + N_{incomp} + N_{neg}, \\
    g_i^x=\Delta x_i,  g_i^w=\Delta w_i,  r_i^x=\Delta \hat{x_i},  r_i^w=\Delta \hat{w_i},\\
    g_i^s=\Delta s_i,  g_i^e=\Delta e_i,  r_i^s=\Delta \hat{s_i},  r_i^e=\Delta \hat{e_i}
    \end{split}
\end{equation}
The final objective function is written as:
\begin{equation}
    Loss = L_{iou} + \lambda L_{reg},
\end{equation}
where $\lambda$ is a balance parameter, we set it to 1.0 empirically.

\begin{table}[t]\footnotesize
\centering
\caption{Comparison with state-of-the-art methods on HACS. The results are measured by mAP(\%) at different tIoU thresholds and average mAP(\%). * indicates our implementation.}
\begin{tabular}{ccccc}
\toprule
Method & 0.5 & 0.75 & 0.95 & Average \\
\midrule
2019-Winner~\cite{zhang2019hacs_winner}  & -     & - & -   &  23.49 \\
BMN~\cite{lin2019bmn}*  &    52.49    &  36.38 &  10.37  &  35.76 \\
\midrule
{ TCANet[SW]}    &  {54.14}    & {37.24} & {11.32}  & {36.79} \\
TCANet[BMN]    &  \textbf{56.74}    & \textbf{41.14} & \textbf{12.15}  & \textbf{39.77} \\
\bottomrule
\end{tabular}
\label{tab:hacs_sota_table}
\vspace{-0.2cm}
\end{table}

\begin{table}[t]\footnotesize
\centering
\caption{Comparison between our TCANet with other state-of-the-arts methods on ActivityNet-v1.3. The results are measured by mAP(\%) at different tIoU thresholds and average mAP(\%). For fair comparisons, we combined our proposals with video-level classification results from ~\cite{xiong2016cuhk}. * indicates the reproduced results.
% that the paper did not report the results with this feature, and the results are reproduced by us.
}
\begin{tabular}{ccccc}
\toprule
Method & 0.5 & 0.75 & 0.95 & Average \\
\midrule
% P-GCN(I3D)  & 48.26     & 33.16 & 3.27   &   31.11 \\
BSN~\cite{lin2018bsn}(2stream)  & 46.45     & 29.96 & 8.02   &   30.03 \\
BMN~\cite{lin2019bmn} (2stream)  &    50.07    &  34.78 &  8.29  &  33.85 \\
G-TAD~\cite{xu2020gtad} (2stream)  & 50.36     & 34.60 & 9.02   &   34.09 \\
BSN++~\cite{su2020bsn++} (2stream)  &    51.27   &  35.70 &  8.33  &  34.88 \\
BMN~\cite{lin2019bmn} (SlowFast)*  &    52.24   &  35.89 &  8.33  &  35.28 \\
% BC-GNN  & 50.56     & 34.75 & 9.37   &   34.26 \\
\midrule
PGCN~\cite{zeng2019pgcn}[BSN] (I3D)  &    48.26  & 33.16  & 3.27   & 31.33 \\
\tabincell{c}{\small TCANet[BSN]  (2stream)} & 51.91  & 34.92  &  7.46  &  34.43 \\
\midrule
TCANet[BMN] (2stream)&  \textbf{52.27}    & \textbf{36.73} & \textbf{6.86}  & \textbf{35.52} \\
TCANet[BMN] (SlowFast)&  \textbf{54.33}    & \textbf{39.13} & \textbf{8.41}  & \textbf{37.56} \\
\bottomrule
\end{tabular}
\vspace{-0.2cm}
\label{tab:anet_sota_table}
\end{table}

\subsection{Inference}
In inference, proposals utilized by TCANet should have a high recall rate.
Therefore, proposals output by BMN~\cite{lin2019bmn} are directly adopted as the input of TCANet.
% , then the refined proposals and score predictions output by the last TBR are the output of TCANet. 
%
The final confidence of proposals are obtained by fusing the BMN score and TCANet score:
\begin{equation}
    S_{proposal} = S_{BMN} * S_{TCANet}
\end{equation}

Finally, Soft-NMS~\cite{bodla2017softnms} is employed to remove redundant proposals.

\section{Experiments}
\subsection{Datasets and Setup}
\noindent \textbf{HACS}~\cite{zhao2019hacs} is a large-scale dataset for temporal action detection. 
It contains 37.6k training, 6k validation, and 6k testing videos with 200 action categories.

\noindent \textbf{ActivityNet-v1.3}~\cite{caba2015activitynet} is a popular benchmark for temporal action detection. It contains 10k training, 5k validation, and 5k testing videos with 200 action categories.

\noindent \textbf{THUMOS14}~\cite{thumos14} contains 200 validation videos and 213 testing videos, including 20 action categories. In our experiments, we compare TCANet with the state-of-the-art method on all three datasets and performed ablation studies on HACS dataset.

\noindent \textbf{Evaluation Metrics.}
Average Recall (AR) is the average recall rate under specified tIoU thresholds for measuring the quality of proposals. On HACS and  ActivityNet-v1.3, these thresholds are set to [0.5:0.05:0.95]. On THUMOS14, they are set to [0.5:0.05:1.0].
By limiting the average number (AN) of proposals for each video , we can calculate the area under the AR vs AN curve to obtain AUC. On  ActivityNet-v1.3, AN is set from 1 to 100.
The quality of temporal action detection requires to evaluate mean Average Precision(mAP) under multiple tIoU.
On HACS and  ActivityNet-v1.3, the tIoU thresholds are set to \{0.5,0.75,0.95\}, and we also test the average mAP of tIoU thresholds between 0.5 and 0.95 with step of 0.05.
On THUMOS14, these tIoU thresholds are set to \{0.3,0.4,0.5,0.6,0.7\}.

\noindent \textbf{Implementation Details.} On HACS and  ActivityNet-v1.3, SlowFast~\cite{feichtenhofer2019slowfast} is adopted to extract a 2304-dimensional feature vector for each snippet. 
Each snippet contains $s=32$ frames and snippet interval $\delta$ is 8. 
For a fair comparison, 2stream network~\cite{wang2015twostream2} is adopted for feature encoding following ~\cite{lin2018bsn, lin2019bmn} on  ActivityNet-v1.3 and THUMOS14.
%
% On THUMOS14, 2stream network is adopted following ~\cite{lin2018bsn, lin2019bmn}.

To reduce information loss, the lengths of the input feature sequence are not down-resized; hence each input sequence is fixed to 1000 and 1500 by zero-padding on HACS and  ActivityNet-v1.3 for batch training, respectively. The Number of groups $N$ and $A$ in LGTE are empirically set to 8 and 4.
The learning rates on these two datasets are set to 0.0004 and 0.001, and the batch size is 16 for 10 epochs.
For THUMOS14 training, a sliding window with a size of 256 is adopted. We set the learning rate, batch size, and epoch number to 0.0004, 16 and 5, respectively.
% of THUMOS14 to 0.0004, the batch size is 16, and the epoch number is 5.

\begin{table}[t]\footnotesize
\centering
\caption{Comparison between our TCANet with other state-of-the-art methods 
% TURN~\cite{gao2017turn}, BSN~\cite{lin2018bsn}, MGG~\cite{liu2019mgg}, BMN~\cite{lin2019bmn}, G-TAD~\cite{xu2020gtad}, BSN++~\cite{su2020bsn++} 
on THUMOS14 dataset. The results are measured by mAP(\%) at different tIoU thresholds. We combined our proposals with video-level classifier UntrimmedNet~\cite{wang2017unet}.}
\setlength{\tabcolsep}{1.4mm}{
\begin{tabular}{ccccccc}
\toprule
Method & Classifier & 0.7 & 0.6 & 0.5 & 0.4 & 0.3 \\
\midrule
TURN~\cite{gao2017turn} & UNet & 6.3     & 14.1 & 24.5   &   35.3 &   46.3 \\
BSN~\cite{lin2018bsn}  & UNet &    20.0    &  28.4 &  36.9  &  45.0 &   53.5 \\
MGG~\cite{liu2019mgg}  & UNet & 21.3  &  29.5   &  37.4   &   46.8   &  53.9 \\
BMN~\cite{lin2019bmn} & UNet &    20.5    &  29.7 &  38.8  &  47.4 &   56.0 \\
G-TAD~\cite{xu2020gtad} & UNet & 23.4     & 30.8 & 40.2   &   47.6 &   54.5\\
BSN++~\cite{su2020bsn++} & UNet & 22.8     & 31.9 & 41.3   &   49.5 &   59.9\\
\midrule
TCANet & UNet &  \textbf{26.7}    & \textbf{36.8} & \textbf{44.6}  & \textbf{53.2} & \textbf{60.6}\\
\bottomrule
\end{tabular}}
\label{tab:thumos_sota_table}
\vspace{-0.2cm}
\end{table}

\begin{table}[t]
\footnotesize
\centering

\caption{Comparison of our TCANet with other state-of-the-art methods
% TAG~\cite{zhao2017tag}, CTAP~\cite{gao2018ctap}, BSN~\cite{lin2018bsn}, MGG~\cite{liu2019mgg}, BMN~\cite{lin2019bmn}, BSN++~\cite{su2020bsn++} 
on THUMOS14 dataset in terms of AR@AN. }
\vspace{0.05cm}
\setlength{\tabcolsep}{1.4mm}{
\begin{tabular}{ccccccc}
\toprule
Feature & Method & @50 & @100 & @200 & @500 & @1000 \\
\midrule
2stream & TAG~\cite{zhao2017tag} & 18.55    & 29.00 & 39.61   &   - &   - \\
2stream & CTAP~\cite{gao2018ctap} & 32.49     & 42.61 & 51.97   &   - &   - \\
2stream  & BSN~\cite{lin2018bsn} &    37.46    &  46.06 &  53.23  &  61.35 &   65.10 \\
2stream & MGG~\cite{liu2019mgg} &    39.93    &  47.75 &  54.65  &  61.36 &   64.06 \\
2stream & BMN~\cite{lin2019bmn} & 39.36     & 47.72 & 54.84   &   62.19 &   65.49\\
2stream & BSN++~\cite{su2020bsn++} & \textbf{42.44}  & 49.84 & \textbf{57.61} &  \textbf{65.17} &  66.83\\
\midrule
2stream & TCANet & 42.05   & \textbf{50.48} & 57.13  & 63.61 & \textbf{66.88}\\
\bottomrule
\end{tabular}}
\label{tab:thumos_auc_sota_table}
\vspace{-0.2cm}
\end{table}

\subsection{Comparison with State-of-the-art Results}
This section will compare with the existing state-of-the-art methods on HACS, ActivityNet-v1.3, and THUMOS14.

\noindent \textbf{HACS. }
On HACS, TCANet is compared with the existing methods in Table~\ref{tab:hacs_sota_table} on the validation set. 
TCANet using only a single model significantly surpass the existing methods. 
Compared with the benchmark method BMN, TCANet's mAP is improved by 4\%.

\noindent \textbf{ActivityNet-v1.3.}
Table~\ref{tab:anet_sota_table} and Table~\ref{tab:anet_auc_sota_table} compare TCANet with other methods, where TCANet significantlys improve both the temporal action proposal and detection performance.
For a fair comparison, TCANet is conducted on the 2stream features for experiments. 
% Notice that TCANet can well correct the candidate proposals, especially when the quality of the candidate proposals is poor.
%
Under the same settings, TCANet can also obtain 1.67\% mAP improvement compared with BMN and significantly outperform other existing methods.

\noindent \textbf{THUMOS14.}
We compare TCANet with the state-of-the-art methods on THUMOS14 in Table~\ref{tab:thumos_sota_table} and Table~\ref{tab:thumos_auc_sota_table}.
% Following~\cite{lin2019bmn}, the proposals obtained by our TCANet are fused with the video-level classification results obtained by UntrimmedNet~\cite{wang2017unet} to generate the action detection results. 
%
% Although the detection performance are more improved than the recall rate, which fully proves that TCANet can generate higher quality proposals.
Since that our TCANet improves the Average Recall with  the  first  several  proposals, the detection performance are more improved than the recall rate.
Especially, in Table~\ref{tab:thumos_sota_table}, when t$Iou$=0.6, TCANet is 4.9\% higher than BSN++~\cite{su2020bsn++}.

\begin{table}[t]\footnotesize
\centering
\caption{Comparison between our TCANet with other state-of-the-art methods  CTAP~\cite{gao2018ctap}, BSN~\cite{lin2018bsn}, MGG~\cite{liu2019mgg}, BMN ~\cite{lin2019bmn} on ActivityNet-v1.3 in terms of AR@AN and AUC.}
\begin{tabular}{cccccc}
\toprule
Method & CTAP & BSN &  MGG & BMN & TCANet \\
\midrule
AR@1(val)  & -  &   32.17   &  -    &  -  & \textbf{34.55}\\
AR@100(val)  & 73.17  &   74.16   &  74.54    &  75.01  & \textbf{76.08}\\
AUC(val)   &  65.72 &   66.17   &  66.43  &   67.10  & \textbf{68.08}\\
\bottomrule
\end{tabular}
\label{tab:anet_auc_sota_table}
\vspace{-0.2cm}
\end{table}
\begin{table}[t]\small
\centering
\caption{Ablation study of TBR, LGTE and progressive refinement strategy on HACS dataset in terms of average mAP(\%).}
\begin{tabular}{ccccc}
\toprule
TBR1 & TBR2 & TBR3        &       LGTE                &  Average\\
\midrule
  &   &      &      &  35.76 \\
\cmark  &   &      &    &   37.16 \\
\cmark  &\cmark  &    & &   37.45 \\
\cmark  &\cmark  &  \cmark   &  &    37.78 \\
\cmark  &\cmark  &  \cmark   &  \cmark  &   \textbf{38.71} \\
\bottomrule
\end{tabular}
\label{tab:hacs_multi_stage}
\vspace{-0.2cm}
\end{table}

\begin{table}[t]\small
\centering
\caption{The effect of different \textit{window size} settings in the local dependency matrix of LGTE on HACS in terms of average mAP(\%).}
\begin{tabular}{ccccc}
\toprule
$Window Size$ & 0.5 & 0.75 & 0.95 & Average \\
\midrule
5  & 55.27 & 39.54  &  11.61  &  38.41 \\
9  & \textbf{55.60}    &  40.01 &  11.47  &  \textbf{38.71} \\
15 &   55.56  & 39.83  &  11.89  &  38.70 \\
25 &   54.99  & \textbf{40.06}  &  \textbf{11.94}  &  38.67 \\
T(GTE only) & 54.70 & 39.64  & 11.71  & 38.37\\
\bottomrule
\end{tabular}
\label{tab:hacs_window_size}
\vspace{-0.2cm}
\end{table}

\subsection{Ablation Study}
In this section, we conduct ablation studies on HACS to verify the effectiveness of each module in TCANet. 

\noindent \textbf{Is progressive refinement necessary?}
The progressive refinement strategy is a part of our TCANet. 
%
% Intuitively, if one stage of refinement can achieve good enough performance, progressive refinement is entirely unnecessary. 
%
Here, the necessity of progressive refinement is illustrated by the separation experiment of three TBRs in Table~\ref{tab:hacs_multi_stage}.
%
% In Table~\ref{tab:hacs_multi_stage}, we discarded three TBRs in turn, showing the results from no refinement to three stages.
%
% Proposals refinement of TCANet is a gradual process, and each TBR has a positive effect on the quality of proposals. 
Although each TBR has a positive effect on performance,
with more stages, this promotion is gradually weakened.
Thus TCANet only contains three stages.

\begin{table}[t]
\centering
\caption{The effect of different groups $N$ LGTE on HACS dataset in terms of average mAP(\%).}
\begin{tabular}{ccccc}
\toprule
$N$ & 0.5 & 0.75 & 0.95 & Average \\
\midrule
2  & 55.27 &  39.86 &  10.91  & 38.49  \\
4  & 55.13    &  39.65 &  11.28  &  38.38 \\
8 &   \textbf{55.60}    &  \textbf{40.01} &  11.47  &  \textbf{38.71} \\
16 &  54.87   & 40.00  & \textbf{11.63} & 38.56  \\
\bottomrule
\end{tabular}
\label{tab:hacs_groups}
\vspace{-0.2cm}
\end{table}

\begin{table}[t]\small
\centering
\caption{The effect of the number of LGTE on HACS dataset in terms of average mAP(\%).}
\begin{tabular}{ccccc}
\toprule
Number of LGTE  & 0.5 & 0.75 & 0.95 & Average \\
\midrule
0  & 54.73    &  39.05 &  10.72  &  37.78 \\
1  &   55.13  &  39.67 &  11.42  &  38.31 \\
2 & 55.60    &  40.01 &  11.47  &  38.71   \\
4 & \textbf{55.72}    &  \textbf{40.03} &  \textbf{11.73}  &  \textbf{38.85} \\
6 &   55.65  & 40.02  &  11.73  &  38.80 \\
\bottomrule
\end{tabular}
      
\label{tab:hacs_LGTE_num}
\vspace{-0.2cm}
\end{table}

\begin{table}[t]\small
\centering
\caption{The generalizability of LGTE under different frameworks on HACS dataset in terms of average mAP(\%).}
\begin{tabular}{cccccc}
\toprule
Framework & LGTE  & 0.5 & 0.75 & 0.95 & Average \\
\midrule
BMN & \xmark & 52.49 &  36.38 &  10.37  &  35.76 \\
BMN & \cmark & 54.75    &  38.72 &  11.41  &  37.76 \\
TCANet & \xmark & 54.73    &  39.05 &  10.72  &  37.78 \\
TCANet & \cmark & \textbf{55.60}    &  \textbf{40.01} &  \textbf{11.47}  &  \textbf{38.71} \\
\bottomrule
\end{tabular}
\label{tab:hacs_different_framework}
\vspace{-0.2cm}
\end{table}

\noindent \textbf{What \textit{WindowSize} and groups  in LGTE should be set?}
In Table~\ref{tab:hacs_window_size} and Table~\ref{tab:hacs_groups}, we conduct experiments to explore the effect of \textit{WindowSize}.
If the \textit{WindowSize} is set extremely small(\textit{WindowSize} = 5 or smaller), the local groups' features fail to collect enough local details. On the contrary(\textit{WindowSize}=T), they will introduce excessive global noise. The number of groups $N$ determines whether various temporal relationships can be modeled.
% The number of N determines whether various temporal relationships can be modeled.
% but also increase the computational cost.
%
% , which shows that a better result can be obtained when the \textit{window size} is set to 9 or 15. 
%
% If the \textit{window size} is set to $T$, it means that there is no LTE, only GTE, which leads to the lack of modeling for local changes.
%
Considering the performance,
% and the computational cost comprehensively, 
we finally set the \textit{WindowSize} to 9 and the groups to 8 in our experiments.

\noindent \textbf{What is the effect of the number of LGTE?}
As an easy-plug-in module, performance can be improved by stacking multiple LGTEs.
%
% In this section, LGTEs are stacked in TCANet. 
% Table~\ref{tab:hacs_LGTE_num} shows the effect of the number of LGTE.
%
% As the number increases, the performance of the TCANet has improved significantly.
Table~\ref{tab:hacs_LGTE_num} demonstrates the performance of the TCANet  improves significantly with the increase of LGTE.
However, excessive LGTE  will lead to over-fitting. 
The performance of TCANet can reach the best with four LGTEs.
Nevertheless, two LGTEs are employed in other ablation studies to facilitate the experiments.

\noindent \textbf{Is LGTE general?}
To validate the generalizability of our proposed LGTE, we also add it to the BMN~\cite{lin2019bmn} framework. 
The experimental results are shown in Table~\ref{tab:hacs_different_framework}, which reveals that LGTE can also significantly improve the performance of BMN and demonstrate the importance of temporal relationship modeling for temporal action localization task.

{\color{black}
\noindent \textbf{How does the frame-level regression affect the TBR?}
To verify the effect of frame-level regression in TBR, we conducted experiments using both regression methods separately, as shown in Table~\ref{tab:hacs_xwse}. 
If only frame-level regression is applied, the detection performance will drop with only boundary local information. 
%
% However, when tIoU is equal to 0.5 or 0.75, the mAP can be improved. 
%
The two methods are combined to boost performance in the final average mAP.}

\begin{figure}[t]
\centering
\includegraphics[width=7.0cm]{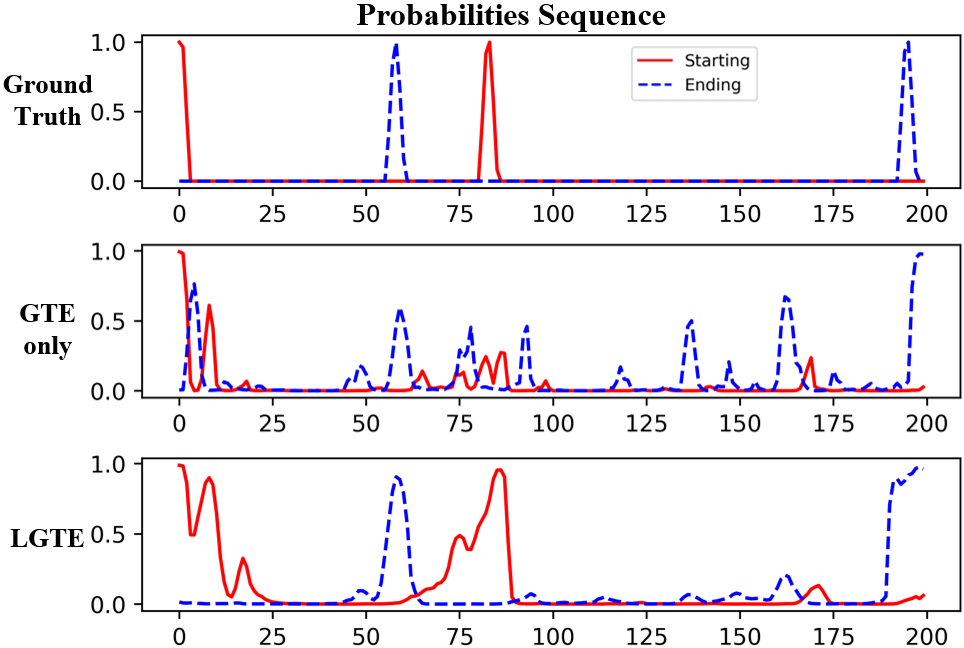}
\caption{
% The effect of LGTE and GTE on the prediction of starting and ending probability sequences. 
The starting and ending probability sequences generated by LGTE and GTE. 
% The figure above is a typical prediction example. 
% From top to bottom, there are three kinds of probability sequences: ground truth, generated by GTE only, and generated by LGTE.
}
\label{fig_pro_seq}
\vspace{-0.2cm}
\end{figure}

\begin{table}[t]\small
\centering
\caption{The effect of Frame-Level Regression on HACS in terms of mAP(\%). SLR and FLR indicate Segment-Level Regression and Frame-Level Regression, respectively.}
\begin{tabular}{cccccc}
\toprule
\makecell[c]{SLR}
 & \makecell[c]{FLR}  & 0.5 & 0.75 & 0.95 & Average \\
\midrule
\cmark   & \xmark  & 54.58 & 39.24 & \textbf{11.72}   &  38.22 \\
\xmark  & \cmark &   55.02  & 39.61  &  9.14  &  37.92 \\
\cmark & \cmark & \textbf{55.60}    &  \textbf{40.01} &  11.47  &  \textbf{38.71} \\
\bottomrule
\end{tabular}

\label{tab:hacs_xwse}
\vspace{-0.2cm}
\end{table}

\noindent \textbf{Why mAP not AUC?}
In our experiments, we find that the detection metric (mAP) mainly depends on Average Recall (AR) with the first several proposals, while the proposal metric (AUC) depends on the first 100 proposals. Hence AR with a small number of proposals has a higher weight in the evaluation metric of detection performance. Extensive experiments have shown that our TCANet can generate fewer proposals with high recall than other methods. Thus the performance improvement of the detection metric is obvious than the proposal metric.

\noindent \textbf{Efficiency Analysis.}
The input candidate proposals for TCANet need to ensure a high recall rate. 
Taking the BMN-generated proposals as an example, when 2000 candidate proposals are selected, the recall rate can reach 91\% with tIOU=0.5.
Our test results are shown in Table~\ref{tab:inference_time}. For a video, LGTE only needs to encode video features once, and the TBR can process multiple proposals in parallel. Therefore, TCANet only takes 20.9 ms to handle a 9-minute video with 2000 candidate proposals. Compared with BMN, the time consumed by TCANet is only 10\%.

\subsection{Visualization}
\begin{figure}[t]
\centering
\includegraphics[width=8.5cm]{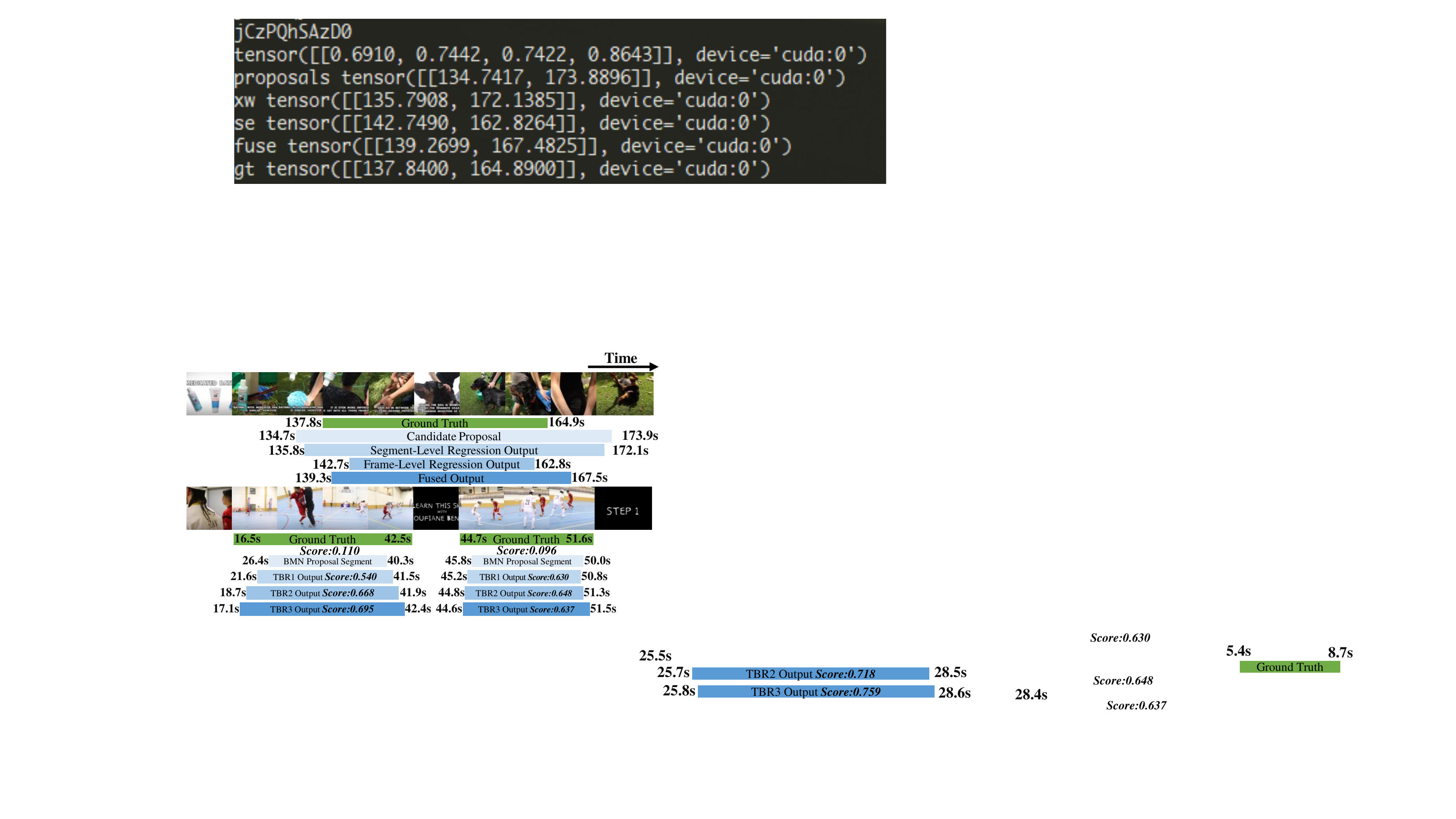}
\caption{Qualitative examples of proposals generated by TBR(top) and TCANet(bottom) on HACS dataset.
}
\label{fig_example}
\vspace{-0.2cm}
\end{figure}

To further explore the interpretability of LGTE, GTE only and LGTE are both leveraged to embed the input video features. To facilitate observation, the obtained features are utilized to predict the starting and ending probability sequences. An example is shown in Figure~\ref{fig_pro_seq}. 
% It is  observed that not only the boundary obtained by LGTE is smoother than that obtained by GTE, but also the boundary location is more accurate. This proves that LGTE can reduce global noise and enhance the ability of boundary awareness.
It is observed that the boundary obtained by LGTE is more accurate and smoother than that obtained by GTE. This indicates that LGTE can reduce global noise and enhance the boundary awareness.
Figure~\ref{fig_example} shows the output of TBR and TCANet. In the top row, both the segment-level and the frame-level output can improve the candidate proposals, but the boundaries are not accurate. The fusion of these two outputs can make the proposal closer to the ground truth.
The bottom row shows that our TCANet can generate the proposals from coarse to fine, and provide more reliable confidence scores, especially for short-term action instances.

\begin{table}[t]\small
\centering
\footnotesize
\caption{The inference time of each module in TCANet on HACS dataset. 2000 candidate proposals were utilized as input to TCANet, and a Nvidia 1080Ti graphic card was employed to process a video for about 9 minutes.}
\begin{tabular}{cccccc}
\toprule
Num$\times$Module & 1$\times$BMN & 2$\times$LGTE & 3$\times$TBR &  Total \\
\midrule
Num$\times$Time Cost& 1$\times$181ms & 2$\times$1.6ms  & 3$\times$5.9ms  &  201.9ms \\
\bottomrule
\end{tabular}

\label{tab:inference_time}
\vspace{-0.2cm}
\end{table}

\section{Conclusion}
In this paper, we propose a novel Temporal Context Aggregation Network (TCANet) for temporal action proposal generation. Firstly we introduce the Local-Global Temporal Encoder (LGTE) to capture both \textit{local and global} temporal relationships simultaneously in a channel grouping fashion. Then the complementary boundary regression mechanism is designed to obtain more precise boundaries and confidence scores. Extensive experiments conducted on several famous benchmarks demonstrate that our TCANet can achieve significant improvement on both action proposal and action detection performance. 

\section{Acknowledgment}
This work is supported by the National Natural Science Foundation of China under grant 61871435 and the Fundamental Research Funds for the Central Universities no. 2019kfyXKJC024.

{\small
\bibliographystyle{ieee_fullname}
\bibliography{egbib}
}

\end{document}